\documentclass[conference]{IEEEtran}
\usepackage[T1]{fontenc}
\usepackage{cite}
\usepackage{amsmath}
\usepackage{graphicx}
\usepackage{array}
\usepackage{booktabs}
\usepackage[table]{xcolor}
\usepackage{url}
\usepackage[hidelinks]{hyperref}
\hypersetup{
  pdftitle={How Prompt Style Affects AI Accuracy and Speed in Competitive Programming},
  pdfauthor={Saad B. Alarifi, Mohammed A. Alam}
}

\newcommand{\twoline}[2]{\begin{tabular}[c]{@{}c@{}}\textbf{#1}\\\textbf{#2}\end{tabular}}

\begin{document}

\title{How Prompt Style Affects AI Accuracy and Speed\\in Competitive Programming}

\author{\IEEEauthorblockN{Saad B. Alarifi}
\IEEEauthorblockA{Riyadh Schools\\
Riyadh, Saudi Arabia\\
saadalarifi2009@gmail.com}
\and
\IEEEauthorblockN{Mohammed A. Alam}
\IEEEauthorblockA{Department of Computer Science\\
Northwestern University\\
Evanston, IL, USA\\
mohammed.alam@northwestern.edu}}

\maketitle

\begin{abstract}
Large language models are increasingly used in competitive programming, and their outputs are sensitive to how a prompt is phrased. Prior work has compared prompt strategies for accuracy on competitive programming, but not for runtime speed, and has not situated the results against human performance. In an exploratory study of one prompt in each of nine styles on GPT-5.5 across three accuracy tasks from the International Olympiad in Informatics (IOI) and two NP-hard speed tasks, we found that no single prompt was best: the strongest prompt depended on the task type and on whether results were summarized by the mean or the median, suggesting that, at least for these tasks, prompt style should be matched to the objective rather than treated as universal. Many differences in mean scores came from a few outlier runs, showing that evaluating prompting by a single metric can obscure how much its effectiveness varies. Compared descriptively with human IOI contestants, who worked under different conditions, the AI was neither strictly better nor strictly worse, exceeding the average contestant on one task but falling short on the other two. By extending prompt comparison from accuracy to runtime speed and grounding it against human performance, this study gives a broader picture of how prompt style affects competitive programming performance.
\end{abstract}

\section{Introduction}

Large language models have become increasingly capable at code generation and competitive programming. At the same time, their outputs are known to be sensitive to how a prompt is phrased; prior work has found that even meaning-preserving changes in prompt formatting can shift model performance substantially~\cite{sclar2023quantifying}. This raises a natural question: does the way a competitive programming task is phrased to an AI change how well it solves the task?

Prior work has examined how prompting techniques affect performance on software engineering tasks~\cite{santana2025prompting}, and has separately measured AI problem-solving ability in competitive programming without comparing prompt styles~\cite{openai2025competitive,zheng2025livecodebenchpro}. Prompt strategies have been compared for accuracy on competitive programming problems~\cite{padiyar2025analyzing}, but not for runtime speed. We address this gap by comparing nine prompt styles across both accuracy and speed. To ground these results, we also situate the AI's accuracy against human contestants at the International Olympiad in Informatics (IOI), an annual competitive programming olympiad for secondary-school students~\cite{ioi}.

In this exploratory study, we evaluated nine prompt styles on five competitive programming tasks: three IOI tasks measuring accuracy and two NP-hard tasks measuring speed~\cite{ioi,ioi2022prisoner,ioi2024message,ioi2025migrations,garey1979}. To account for run-to-run variability, each prompt style was run nine times per task. This paper makes three contributions.

First, it compares prompt styles on speed, measured as the largest input a solution can handle within a time limit, as well as on accuracy. No single prompt style performed best across all tasks: the strongest style depended both on the task type (accuracy versus speed) and on the metric used to summarize it (mean versus median). The ``best prompt'' is therefore not a stable property but one that depends on how performance is measured.

Second, it examines how much of the difference between prompt styles is consistent and how much comes from individual runs. After correcting for testing five tasks, prompt style had a statistically significant effect on only two of them, and many differences in mean scores were driven by a minority of outlier runs rather than consistent effects. Comparisons of prompts based on a single summary score can therefore be misleading.

Third, it situates the AI's accuracy against human IOI contestants as descriptive context. The AI was neither strictly better nor strictly worse: it exceeded the average contestant on one task while falling short on the other two, and remained below gold-medalist level throughout. How the AI compares to humans therefore depends heavily on which task is used.

All prompts, generated solutions, grading scripts, and speed test cases are publicly released, so the study can be re-analyzed or extended with more prompts, runs, and models.

\section{Related Work}

\subsection{AI in Competitive Programming}

AlphaCode showed that AI can be used to solve competitive programming tasks by achieving an average ranking of top 54.3\% in competitions with more than 5,000 participants~\cite{li2022alphacode}. More recent large reasoning models have reported stronger results, with OpenAI achieving gold-medal performance at IOI 2024~\cite{openai2025competitive}. In contrast, a benchmark study done by Olympiad medalists found that frontier models solved none of the hard problems that human experts can solve~\cite{zheng2025livecodebenchpro}. Benchmarks have also compared LLMs directly against human contestants on Olympiad problems: LiveOIBench evaluates models on subtask-graded Olympiad tasks against elite contestant percentiles, finding that even frontier models fall short of top human contestants~\cite{zou2025liveoibench}. These works compare models against humans but do not vary the prompt.

\subsection{Prompting Techniques for Code Generation}

Several prompting techniques have been proposed specifically for code generation. Self-planning code generation has the model first outline a solution plan before implementing it~\cite{jiang2023selfplanning}, and Structured Chain-of-Thought prompting has the model reason in terms of program structures prior to writing code~\cite{li2023scot}. More broadly, empirical comparisons of prompting techniques have been done for software engineering tasks~\cite{santana2025prompting}. Prompt effects on the runtime efficiency of generated code have also been examined on standard coding benchmarks such as HumanEval and LeetCode, where prompting for efficiency yielded only limited gains~\cite{qiu2025efficient,niu2024evaluating}.

\subsection{Our Contribution}

The closest prior work to ours compares prompt strategies for accuracy on competitive programming problems~\cite{padiyar2025analyzing}, but stops at accuracy. Separately, prompt effects on runtime efficiency have been studied on standard coding benchmarks~\cite{qiu2025efficient,niu2024evaluating}, and LLMs have been benchmarked against human contestants on Olympiad problems without varying the prompt~\cite{zheng2025livecodebenchpro,zou2025liveoibench}. Our work extends the accuracy-focused prompt comparison to runtime efficiency, which has not been measured for prompt styles in this setting. We additionally contextualize the AI's accuracy results against human IOI contestants, giving a reference point for how the prompt styles perform in absolute terms. Together this gives a broader picture of how prompt style affects competitive programming performance than accuracy alone.

\section{Data}

\subsection{Benchmark Selection}

Five competitive programming tasks were used as benchmarks, three of which were used to measure accuracy and two of which were used to measure speed.

\subsection{Accuracy Benchmarks}

The three tasks used to measure accuracy were taken from the IOI~\cite{ioi}:
\begin{itemize}
\item \textbf{Prisoner Challenge (IOI 2022)}~\cite{ioi2022prisoner}: design a strategy for up to 500 prisoners who, one at a time, inspect one of two bags of coins and communicate only by rewriting a single number on a whiteboard, so that one of them identifies the bag with fewer coins; the score depends on the largest number the strategy needs to write.
\item \textbf{Message (IOI 2024)}~\cite{ioi2024message}: send a message of bits through 31-bit packets, 15 bits of which can be changed by an adversary at positions known to the sender but not the receiver, so that the receiver can reconstruct the message exactly; the score depends on the number of packets sent.
\item \textbf{Migrations (IOI 2025)}~\cite{ioi2025migrations}: as a tree is revealed one vertex at a time, send at most 50 messages, each an integer from 1 to 20,000, from which a receiver must identify two vertices that are the farthest apart; the score depends on the largest integer sent and the number of messages.
\end{itemize}

Each of these tasks has a list of subtasks in which it is possible to gather partial points, so solutions of different quality receive different scores. These tasks were picked because this subtask-based scoring makes it possible to compare solution quality more finely. Furthermore, their relatively low constraints make time-limit issues less significant, allowing these benchmarks to isolate accuracy; the same low constraints, however, make them unsuitable for measuring speed, which is why the two speed tasks are evaluated separately.

\subsection{Speed Benchmarks}

The two tasks used to measure speed are Maximum Clique Size, which asks for the size of the largest group of vertices in a graph that are all connected to one another, and Number Partitioning, which asks for the smallest possible difference between the sums of two groups when $N$ positive integers are split between them. Both are NP-hard problems~\cite{garey1979}. These problems are computationally difficult optimization benchmarks, which makes them useful for observing differences in runtime across prompt styles.

\section{Methodology}

\subsection{Prompt Styles}

Nine prompt styles were evaluated across all tasks. Each style is listed below with the main idea of what it asks for; the full style instructions are in Appendix~\ref{app:styles}.
\begin{itemize}
\item \textbf{Prompt 1: Minimal lossless compression style.} As short as possible while keeping every detail the solver needs.
\item \textbf{Prompt 2: Official statement style.} A formal, numbered specification with clear sections.
\item \textbf{Prompt 3: Symbol-definition style.} Defines every symbol and variable first, then restates the task formally.
\item \textbf{Prompt 4: Imperative directive style.} Direct commands to the solver with strong instruction wording.
\item \textbf{Prompt 5: Structured checklist style.} The task as a compact checklist of requirements and constraints.
\item \textbf{Prompt 6: Competitive benchmark style.} Frames the task as a competition against other AI solvers to produce the strongest solution.
\item \textbf{Prompt 7: Educational teaching style.} Explains the task as if to a student seeing it for the first time, with intuition before formal details.
\item \textbf{Prompt 8: Plan-first algorithm design style.} Has the solver plan the algorithm, data structures, and target complexity internally before coding.
\item \textbf{Prompt 9: Structured computational reasoning style.} Has the solver lay out the computation as stages and decision rules internally before coding.
\end{itemize}

Two of the nine prompt styles draw directly on prior work. Prompt 8 was inspired by self-planning code generation~\cite{jiang2023selfplanning}, and Prompt 9 by Structured Chain-of-Thought prompting~\cite{li2023scot}. In both cases, the technique was adapted to a single-shot setting, where planning and structured reasoning are performed internally rather than emitted as intermediate output.

As an example, two of the nine prompts generated for Number Partitioning are shown below; all nine are listed in Appendix~\ref{app:prompts}.

\noindent\textbf{Prompt 1 (minimal lossless compression style):}
\begin{quote}\small
Given N positive integers (line 1: N; next N lines: Ai), split all numbers into two groups minimizing |sum(group1)-sum(group2)|. Print that minimum difference. Example input: 4 1 5 7 9; output: 0. Tests run by increasing N and stop at first failure; TLE gives partial credit up to last solved N, WA/RTE gives zero. Time limit 1s. Use only the C++ standard library. Output only valid C++20 code, no explanation, no markdown.
\end{quote}

\noindent\textbf{Prompt 6 (competitive benchmark style):}
\begin{quote}\small
You are competing against other solvers on an exact two-way partition benchmark. Given N positive integers from stdin, split them into two groups so the absolute difference of group sums is as small as possible, then print that minimum difference. Input is N on the first line followed by N lines containing Ai. Example: 4 / 1 / 5 / 7 / 9 must print 0. Time limit is 1 second, C++20 standard library only. Tests are ordered by increasing N; a TLE stops the run and earns credit up to the largest solved N, while any WA or RTE scores zero, so exactness and robustness are mandatory. Produce the strongest exact implementation you can for large N under these rules. Output only valid C++20 code, with no explanation or markdown.
\end{quote}

\subsection{Experiments}

\subsubsection{Prompt generation}
For each task, GPT-5.5 was given the task statement and the style instructions in Appendix~\ref{app:styles}, and asked to write one prompt in each style. All nine prompts for a task were generated with the reasoning effort set to high, giving 45 prompts in total ($9$ prompt styles $\times$ $5$ tasks). Each style is therefore represented by a single prompt per task, so comparisons between styles are, strictly, comparisons between these particular prompts. The model was instructed to keep each prompt fully self-contained, to never mention contest names, task names, or sources, and to instruct the solver to only output valid C++20 code with no markdown, explanation, or code fences.

\subsubsection{Solution generation}
Each of the 45 prompts was then sent to GPT-5.5, with the reasoning effort set to medium, nine times. This produced nine solutions for each prompt style on each task, or 405 solutions in total ($9$ prompt styles $\times$ $5$ tasks $\times$ $9$ iterations). Every request was a separate API call with no conversation history, so no iteration could see the output of another. The same instruction to output only valid C++20 code, with no markdown, explanation, or code fences, was appended to every prompt, and any code fences left in a response were removed before compiling. Reasoning effort was not set higher due to the long time taken to generate a response.

\subsubsection{Accuracy evaluation}
For the accuracy tasks, the code generated by the prompts was run through the oj.uz online judge platform~\cite{ojuz}. These solutions were given points based on a task-specific grading system, which was included in the prompt given to the AI solver to prevent it from optimizing for the wrong objective. The test cases and grading were provided by the oj.uz platform~\cite{ojuz}, using the official IOI scoring defined for each task~\cite{ioi2022prisoner,ioi2024message,ioi2025migrations}.

\subsubsection{Speed evaluation}
The speed tasks were run locally, with the settings listed in Table~\ref{tab:speed}. Each solution was run on a sequence of test cases of increasing input size $N$, starting at $N = 1$, with one test case per value of $N$, and every solution was evaluated on identical inputs. Solutions were graded one after another in a fixed order, by prompt style and then by iteration. A baseline solution was run on the same test cases with a 10-second limit per case. Its purpose was not to compete on speed but to provide the correct answer for each test case and to establish a ceiling high enough that stronger and weaker solutions would not tie by completing every case. Each AI solution ran until it exceeded the 1-second limit, and the largest $N$ it completed was recorded and normalized to a score out of 100. A wrong answer, runtime error, or compilation error scored zero, so that incorrect but fast solutions were not treated as better-performing than correct solutions. As with the accuracy tasks, the prompts included this grading procedure so that the solver would not optimize for the wrong objective.

\begin{table*}[t]
\centering
\caption{Speed evaluation settings.}
\label{tab:speed}
\small
\renewcommand{\arraystretch}{1.25}
\begin{tabular}{p{0.25\textwidth}p{0.33\textwidth}p{0.33\textwidth}}
\toprule
\textbf{Setting} & \textbf{Maximum Clique Size} & \textbf{Number Partitioning} \\
\midrule
Test input & Random graph on $N$ vertices; each edge included independently with probability $N^{-1/N}$ & $N$ random integers between 1 and $2^{48}$ \\
Test generation & Deterministic from a fixed seed & Deterministic from a fixed seed \\
Baseline solution & MaxCliqueDyn exact algorithm~\cite{konc2007maxclique} & Our own exact meet-in-the-middle solution \\
Baseline time limit & 10\,s per test case & 10\,s per test case \\
Ceiling (largest $N$ solved by baseline) & 220 & 49 \\
AI solution time limit & 1\,s per test case & 1\,s per test case \\
Score & Largest $N$ completed $\div$ 220 $\times$ 100 & Largest $N$ completed $\div$ 49 $\times$ 100 \\
Wrong answer, runtime error, or compilation error & Score of 0 & Score of 0 \\
Compiler & GCC 16.1.0 & Apple Clang 17.0.0 \\
Compiler flags & \texttt{-O3 -march=native -std=c++20} & \texttt{-O3 -march=native -std=c++20} \\
Hardware & 2022 MacBook Air (M2) & 2022 MacBook Air (M2) \\
\bottomrule
\end{tabular}
\end{table*}

\subsubsection{Statistical analysis}
To test whether prompt style had a consistent effect on a task, a Kruskal-Wallis H test~\cite{kruskal1952ranks} was run for each task, comparing the nine prompt styles (nine runs each). The test compares groups by the ranks of their scores rather than the scores themselves, so it does not assume normally distributed scores, and a few extreme runs have little influence on it. This suits the scores in this study, which are heavily tied and contain outliers. To test whether prompt styles that did well on accuracy also did well on speed, the Spearman rank correlation~\cite{spearman1904proof} was computed between each prompt style's average accuracy score and its average speed score, both based on means. All tests used a significance level of 0.05 and were computed with SciPy~\cite{virtanen2020scipy}. Because a separate Kruskal-Wallis test was run for each of the five tasks, the $p$-values were also adjusted with the Holm--Bonferroni method~\cite{holm1979}. As an effect size, we report $\varepsilon^2 = H/(n-1)$, where $n = 81$ is the number of runs per task; it ranges from 0 (no effect) to 1.

\subsection{Human Baseline}

Human scores for the accuracy tasks were taken from the official IOI results for each year~\cite{ioi2022results,ioi2024results,ioi2025results}. For each task, two reference points were computed: the mean score of all contestants, and the mean score of gold medalists, who are roughly the top twelfth of contestants. Both are on the same 0--100 scale as the AI's scores, since the AI's solutions were graded with the official IOI scoring. However, the humans and the AI earned their scores under different conditions (Table~\ref{tab:humans}), which we return to in the Results and Analysis section.

\subsection{Data Recording}
\label{sec:data}

Results and solver code were saved independently for each prompt in each task for each iteration. The full set of prompts, the generated solution and its output for every iteration, the baseline solutions, and the grading scripts, which regenerate the speed test cases from their fixed seeds, are publicly available.\footnote{\url{https://github.com/erering/AI-prompt-research}}

\begin{table*}[t]
\centering
\caption{Scores and conditions for human IOI contestants and the AI solver on the accuracy tasks.}
\label{tab:humans}
\small
\renewcommand{\arraystretch}{1.25}
\begin{tabular}{p{0.20\textwidth}p{0.42\textwidth}p{0.29\textwidth}}
\toprule
 & \textbf{IOI contestants} & \textbf{AI solver (this study)} \\
\midrule
Source of scores & Official IOI results~\cite{ioi2022results,ioi2024results,ioi2025results} & oj.uz, official IOI scoring~\cite{ojuz} \\
Scores averaged per task & Prisoner Challenge (2022): 353 contestants, 30 gold medalists\newline Message (2024): 366 contestants, 34 gold medalists\newline Migrations (2025): 334 contestants, 28 gold medalists & 81 runs (9 prompt styles $\times$ 9 runs) \\
Time & 5 hours per contest day, shared across three tasks & One response per run \\
Submissions per task & Up to 50 & 1 per run \\
Feedback before resubmitting & Subtask results after each submission & None \\
Task score & Sum over subtasks of the best result across all submissions & Score of the single submission \\
\bottomrule
\end{tabular}
\end{table*}

\section{Results \& Analysis}

\subsection{Overview}

\begin{figure}[t]
\centering
\includegraphics[width=\columnwidth]{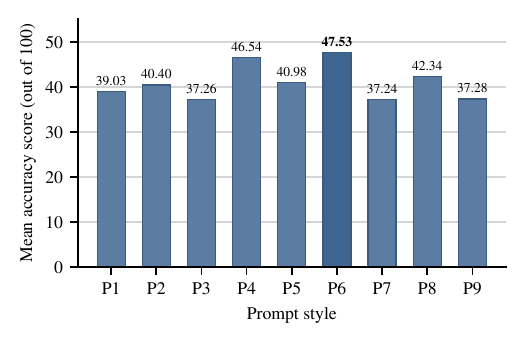}
\caption{Mean accuracy by prompt style, scores are out of 100 and averaged across the three IOI tasks (Prisoner Challenge, Message, Migrations).}
\label{fig:acc}
\end{figure}

Nine prompt styles were tested for accuracy across three IOI tasks and for speed across two NP-hard tasks. Each prompt style had nine iterations for each task giving a total of ($9$ prompt styles $\times$ $5$ tasks $\times$ $9$ iterations) $= 405$ total runs. The following sections report accuracy scores for the three IOI tasks (Tables~\ref{tab:acc-mean}--\ref{tab:acc-median}, Fig.~\ref{fig:acc}), test whether prompt style had a consistent effect on each task (Table~\ref{tab:kw}), report speed scores for the two NP-hard optimization tasks (Tables~\ref{tab:speed-mean}--\ref{tab:speed-median}, Fig.~\ref{fig:speed}), and compare AI performance against human IOI contestants (Fig.~\ref{fig:humans}).

\begin{table}[t]
\centering
\caption{AI accuracy scores by prompt style, scores are out of 100. Mean over 9 iterations per prompt--task pair. Shading compares prompts within each column (green = best, red = worst); bold = column best.}
\label{tab:acc-mean}
\footnotesize
\setlength{\tabcolsep}{3.2pt}
\begin{tabular}{lcccc}
\toprule
\textbf{Prompt} & \twoline{Prisoner}{Challenge} & \textbf{Message} & \textbf{Migrations} & \textbf{Average} \\
\midrule
Prompt 1 & \cellcolor[RGB]{238,203,181}62.78 & \cellcolor[RGB]{237,202,180}30.63 & \cellcolor[RGB]{240,212,183}23.68 & \cellcolor[RGB]{236,197,179}39.03 \\
Prompt 2 & \cellcolor[RGB]{242,229,188}68.89 & \cellcolor[RGB]{234,189,176}29.61 & \cellcolor[RGB]{238,205,181}22.70 & \cellcolor[RGB]{239,210,183}40.40 \\
Prompt 3 & \cellcolor[RGB]{238,205,181}63.17 & \cellcolor[RGB]{235,193,178}29.94 & \cellcolor[RGB]{232,180,174}18.68 & \cellcolor[RGB]{232,180,174}37.26 \\
Prompt 4 & \cellcolor[RGB]{207,220,181}75.00 & \cellcolor[RGB]{237,200,180}30.45 & \cellcolor[RGB]{181,213,176}\textbf{34.16} & \cellcolor[RGB]{193,216,178}46.54 \\
Prompt 5 & \cellcolor[RGB]{243,224,187}67.44 & \cellcolor[RGB]{232,180,174}28.96 & \cellcolor[RGB]{243,229,188}26.55 & \cellcolor[RGB]{241,216,184}40.98 \\
Prompt 6 & \cellcolor[RGB]{181,213,176}\textbf{79.44} & \cellcolor[RGB]{181,213,176}\textbf{36.26} & \cellcolor[RGB]{240,228,187}26.88 & \cellcolor[RGB]{181,213,176}\textbf{47.53} \\
Prompt 7 & \cellcolor[RGB]{232,180,174}57.72 & \cellcolor[RGB]{239,208,182}31.08 & \cellcolor[RGB]{239,207,182}22.93 & \cellcolor[RGB]{232,180,174}37.24 \\
Prompt 8 & \cellcolor[RGB]{242,229,188}68.89 & \cellcolor[RGB]{240,215,184}31.53 & \cellcolor[RGB]{243,229,188}26.59 & \cellcolor[RGB]{244,229,188}42.34 \\
Prompt 9 & \cellcolor[RGB]{235,190,177}60.00 & \cellcolor[RGB]{236,194,178}30.03 & \cellcolor[RGB]{237,200,180}21.81 & \cellcolor[RGB]{232,180,174}37.28 \\
\bottomrule
\end{tabular}
\end{table}

\begin{table}[t]
\centering
\caption{AI accuracy scores by prompt style, scores are out of 100. Median over 9 iterations per prompt--task pair. Shading compares prompts within each column (green = best, red = worst); bold = column best.}
\label{tab:acc-median}
\footnotesize
\setlength{\tabcolsep}{3.2pt}
\begin{tabular}{lcccc}
\toprule
\textbf{Prompt} & \twoline{Prisoner}{Challenge} & \textbf{Message} & \textbf{Migrations} & \textbf{Average} \\
\midrule
Prompt 1 & \cellcolor[RGB]{232,180,174}65.00 & \cellcolor[RGB]{181,213,176}\textbf{31.40} & \cellcolor[RGB]{232,180,174}20.00 & \cellcolor[RGB]{235,192,177}38.80 \\
Prompt 2 & \cellcolor[RGB]{232,180,174}65.00 & \cellcolor[RGB]{244,228,188}30.33 & \cellcolor[RGB]{232,180,174}20.00 & \cellcolor[RGB]{233,186,176}38.44 \\
Prompt 3 & \cellcolor[RGB]{232,180,174}65.00 & \cellcolor[RGB]{232,180,174}29.32 & \cellcolor[RGB]{232,180,174}20.00 & \cellcolor[RGB]{232,180,174}38.11 \\
Prompt 4 & \cellcolor[RGB]{232,180,174}65.00 & \cellcolor[RGB]{181,213,176}\textbf{31.40} & \cellcolor[RGB]{181,213,176}\textbf{28.10} & \cellcolor[RGB]{232,226,186}41.50 \\
Prompt 5 & \cellcolor[RGB]{232,180,174}65.00 & \cellcolor[RGB]{232,180,174}29.32 & \cellcolor[RGB]{232,180,174}20.01 & \cellcolor[RGB]{232,180,174}38.11 \\
Prompt 6 & \cellcolor[RGB]{181,213,176}\textbf{80.00} & \cellcolor[RGB]{181,213,176}\textbf{31.40} & \cellcolor[RGB]{232,180,174}20.00 & \cellcolor[RGB]{181,213,176}\textbf{43.80} \\
Prompt 7 & \cellcolor[RGB]{232,180,174}65.00 & \cellcolor[RGB]{181,213,176}\textbf{31.40} & \cellcolor[RGB]{232,180,174}20.00 & \cellcolor[RGB]{235,192,177}38.80 \\
Prompt 8 & \cellcolor[RGB]{232,180,174}65.00 & \cellcolor[RGB]{181,213,176}\textbf{31.40} & \cellcolor[RGB]{232,180,174}20.01 & \cellcolor[RGB]{235,192,177}38.80 \\
Prompt 9 & \cellcolor[RGB]{232,180,174}65.00 & \cellcolor[RGB]{232,180,174}29.32 & \cellcolor[RGB]{232,180,174}20.00 & \cellcolor[RGB]{232,180,174}38.11 \\
\bottomrule
\end{tabular}
\end{table}

\subsection{Accuracy Results}

While no prompt dominated across all three accuracy tasks, Prompt 6 came the closest. It had the highest mean on both Prisoner Challenge and Message, the highest median on Prisoner Challenge, and tied for the highest median on Message (Tables~\ref{tab:acc-mean}--\ref{tab:acc-median}, Fig.~\ref{fig:acc}). Prompt style also had a significant effect on both of these tasks before correction, and on Prisoner Challenge after correction (Table~\ref{tab:kw}). Because Prompt 6 led on both the mean and the median, its advantage is less likely to come from a few lucky runs, and it may reflect something about the competitive framing itself. On Migrations, however, Prompt 6's median of 20 matched most other prompts, and its second-highest mean there came from just two high-scoring runs. Any advantage from Prompt 6's style therefore appears to depend on the task rather than being universal.

Prompt 4 achieved the highest mean score on Migrations (Table~\ref{tab:acc-mean}) and was the only prompt whose median score was above 20.01 (Table~\ref{tab:acc-median}). Five of its nine runs scored above 20.01, compared with 23 of the 72 runs from the other eight prompts combined. However, prompt style had no significant effect on Migrations overall (Table~\ref{tab:kw}), so this lead may be due to chance rather than to the prompt style. Prompt 4 was also less dominant on the other tasks, performing moderately on Message and strongly, though not best, on Prisoner Challenge. Its result on Migrations is therefore a possibility worth testing with more runs rather than an established effect.

Most prompts tied on median scores (Table~\ref{tab:acc-median}), so much of the spread in the means comes from a minority of unusually high- or low-scoring iterations. The Kruskal-Wallis test, which is little affected by such iterations, gives a more precise picture (Table~\ref{tab:kw}). Prompt style had a significant effect on Prisoner Challenge ($p = 0.012$) and Message ($p = 0.029$), but not on Migrations ($p = 0.53$); after Holm adjustment, only Prisoner Challenge remains significant ($p = 0.048$). The differences between prompts on Migrations are therefore consistent with chance, while on Prisoner Challenge, and less certainly on Message, prompt style appears to shift performance in a way that outlier runs alone do not explain.

\begin{table}[t]
\centering
\caption{Kruskal-Wallis H test of whether prompt style affected scores on each task (9 prompt styles $\times$ 9 runs, 8 degrees of freedom), with Holm-adjusted $p$-values and effect size $\varepsilon^2$. Shading marks significance at the 0.05 level after Holm adjustment (green = significant, red = not significant); bold = significant before adjustment.}
\label{tab:kw}
\small
\setlength{\tabcolsep}{4pt}
\begin{tabular}{llcccc}
\toprule
\textbf{Task} & \textbf{Type} & \textbf{H} & $p$ & \textbf{Holm} $p$ & $\varepsilon^2$ \\
\midrule
Prisoner Challenge & Accuracy & 19.58 & \textbf{0.012} & \cellcolor[RGB]{181,213,176}0.048 & 0.24 \\
Message & Accuracy & 17.16 & \textbf{0.029} & \cellcolor[RGB]{232,180,174}0.086 & 0.21 \\
Migrations & Accuracy & 7.07 & 0.529 & \cellcolor[RGB]{232,180,174}1.000 & 0.09 \\
Maximum Clique Size & Speed & 2.35 & 0.968 & \cellcolor[RGB]{232,180,174}1.000 & 0.03 \\
Number Partitioning$^\dagger$ & Speed & 20.77 & \textbf{0.008} & \cellcolor[RGB]{181,213,176}0.039 & 0.26 \\
\bottomrule
\multicolumn{6}{l}{\footnotesize $^\dagger$Confounded with grading order; see Section~\ref{sec:speed-results}.}
\end{tabular}
\end{table}

\begin{table}[t]
\centering
\caption{AI speed scores by prompt style, normalized to 100. Maximum completed $N$ within the 1\,s time limit, normalized per task; mean over 9 iterations. Shading compares prompts within each column; bold = column best.}
\label{tab:speed-mean}
\footnotesize
\setlength{\tabcolsep}{3.2pt}
\begin{tabular}{lccc}
\toprule
\textbf{Prompt} & \twoline{Maximum}{Clique Size} & \twoline{Number}{Partitioning} & \textbf{Average} \\
\midrule
Prompt 1 & \cellcolor[RGB]{237,227,187}65.35 & \cellcolor[RGB]{236,198,179}80.27 & \cellcolor[RGB]{242,221,186}72.81 \\
Prompt 2 & \cellcolor[RGB]{181,213,176}\textbf{72.32} & \cellcolor[RGB]{234,190,177}79.59 & \cellcolor[RGB]{213,221,182}75.96 \\
Prompt 3 & \cellcolor[RGB]{202,218,180}69.65 & \cellcolor[RGB]{184,214,177}87.30 & \cellcolor[RGB]{181,213,176}\textbf{78.47} \\
Prompt 4 & \cellcolor[RGB]{235,193,178}58.74 & \cellcolor[RGB]{232,180,174}78.68 & \cellcolor[RGB]{232,180,174}68.71 \\
Prompt 5 & \cellcolor[RGB]{239,208,182}61.16 & \cellcolor[RGB]{233,183,175}78.91 & \cellcolor[RGB]{235,193,178}70.04 \\
Prompt 6 & \cellcolor[RGB]{240,214,184}62.02 & \cellcolor[RGB]{223,224,184}84.58 & \cellcolor[RGB]{243,226,187}73.30 \\
Prompt 7 & \cellcolor[RGB]{232,180,174}56.67 & \cellcolor[RGB]{181,213,176}\textbf{87.53} & \cellcolor[RGB]{240,214,184}72.10 \\
Prompt 8 & \cellcolor[RGB]{244,229,188}64.55 & \cellcolor[RGB]{197,217,179}86.39 & \cellcolor[RGB]{220,223,183}75.47 \\
Prompt 9 & \cellcolor[RGB]{237,199,180}59.75 & \cellcolor[RGB]{200,218,180}86.17 & \cellcolor[RGB]{242,223,186}72.96 \\
\bottomrule
\end{tabular}
\end{table}

\subsection{Speed Results}
\label{sec:speed-results}

\begin{figure}[t]
\centering
\includegraphics[width=\columnwidth]{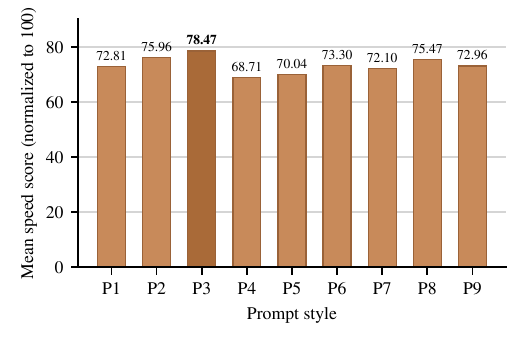}
\caption{Mean speed performance by prompt style, normalized to 100 per task and averaged across the two NP-hard optimization tasks (Maximum Clique Size, Number Partitioning).}
\label{fig:speed}
\end{figure}

While Prompt 3 ranks highest by the mean averaged across the two optimization tasks (Table~\ref{tab:speed-mean}, Fig.~\ref{fig:speed}), it is tied for third on the median for Maximum Clique Size and tied for second for Number Partitioning (Table~\ref{tab:speed-median}). The best-performing prompt on speed therefore depends on whether the measure is the mean or the median.

On Maximum Clique Size, prompt style had no significant effect ($p = 0.97$, Table~\ref{tab:kw}). The variation within each prompt style was far larger than the variation between them: every prompt style had at least one run that completed only up to $N = 87$ or less, and at least one that completed $N = 183$ or more. The ranking of prompts on this task is therefore consistent with run-to-run variation rather than with differences between prompt styles.

Prompt 3 is one of only two prompts that did not score a zero on any iteration across the two optimization tasks. Its high mean relative to its middling median likely comes from avoiding the low-scoring outlier iterations that other prompt styles produced. Failure avoidance alone does not guarantee a high mean, however: Prompt 6 likewise scored no zeros, yet ranked fourth on the mean because its typical runs were slower. Because a single wrong answer or runtime error scores zero, the mean on these tasks rewards reliability as much as raw speed, and Prompt 3's lead on the mean reflects that it never failed more than that it was the fastest.

\begin{table}[t]
\centering
\caption{AI speed scores by prompt style, normalized to 100. Maximum completed $N$ within the 1\,s time limit, normalized per task; median over 9 iterations. Shading compares prompts within each column; bold = column best.}
\label{tab:speed-median}
\footnotesize
\setlength{\tabcolsep}{3.2pt}
\begin{tabular}{lccc}
\toprule
\textbf{Prompt} & \twoline{Maximum}{Clique Size} & \twoline{Number}{Partitioning} & \textbf{Average} \\
\midrule
Prompt 1 & \cellcolor[RGB]{242,221,186}67.27 & \cellcolor[RGB]{181,213,176}\textbf{91.84} & \cellcolor[RGB]{231,226,186}79.56 \\
Prompt 2 & \cellcolor[RGB]{181,213,176}\textbf{80.00} & \cellcolor[RGB]{240,213,183}87.76 & \cellcolor[RGB]{181,213,176}\textbf{83.88} \\
Prompt 3 & \cellcolor[RGB]{242,221,186}67.27 & \cellcolor[RGB]{240,213,183}87.76 & \cellcolor[RGB]{242,221,186}77.52 \\
Prompt 4 & \cellcolor[RGB]{232,180,174}58.18 & \cellcolor[RGB]{240,213,183}87.76 & \cellcolor[RGB]{232,180,174}72.97 \\
Prompt 5 & \cellcolor[RGB]{242,221,186}67.27 & \cellcolor[RGB]{240,213,183}87.76 & \cellcolor[RGB]{242,221,186}77.52 \\
Prompt 6 & \cellcolor[RGB]{232,180,174}58.18 & \cellcolor[RGB]{240,213,183}87.76 & \cellcolor[RGB]{232,180,174}72.97 \\
Prompt 7 & \cellcolor[RGB]{241,217,184}66.36 & \cellcolor[RGB]{240,213,183}87.76 & \cellcolor[RGB]{241,217,184}77.06 \\
Prompt 8 & \cellcolor[RGB]{181,213,176}\textbf{80.00} & \cellcolor[RGB]{232,180,174}85.71 & \cellcolor[RGB]{193,216,178}82.85 \\
Prompt 9 & \cellcolor[RGB]{241,215,184}65.91 & \cellcolor[RGB]{232,180,174}85.71 & \cellcolor[RGB]{238,206,181}75.81 \\
\bottomrule
\end{tabular}
\end{table}

The median results for Number Partitioning span only three values, ranging narrowly from 85.71 to 91.84 (Table~\ref{tab:speed-median}), with six of the nine prompts sharing a median of 87.76. In terms of input size, the typical run of every prompt style completed between $N = 42$ and $N = 45$. These differences are larger than they look: for an exact meet-in-the-middle algorithm like the baseline, each additional step of $N$ multiplies the work by about $\sqrt{2}$, so $N = 45$ requires roughly 2.8 times the work of $N = 42$. The Kruskal-Wallis test found a significant effect of prompt style on this task ($p = 0.008$, Holm-adjusted $p = 0.039$, Table~\ref{tab:kw}): Prompts 8 and 9 never completed more than $N = 43$, while Prompt 1 reached $N = 45$ in five of its nine runs. Because solutions were graded in a fixed order on a fanless laptop, and scores on this task also declined with grading position (Spearman $\rho = -0.61$ among non-zero runs, with Prompts 8 and 9 graded last), part of this difference may reflect thermal throttling, so we treat it as tentative; Maximum Clique Size shows no such trend ($\rho = -0.01$). The larger gaps in the means come instead from single failed runs, as Prompts 1, 2, 4, and 5 each had one iteration that scored zero, so the ranking by mean was decided mostly by which prompts avoided a failed run.

\subsection{Consistency Across Task Types}

The prompts that performed best on accuracy did not perform best on speed. Prompt 6 leads accuracy in both mean and median, while on speed no single prompt leads throughout: Prompt 3 tops the mean, and the median leaders differ by task (Prompts 2 and 8 on Maximum Clique, Prompt 1 on Number Partitioning). In addition, Prompt 4 is second best in average accuracy mean while being last in average speed mean. Across all nine prompt styles, the Spearman rank correlation between average accuracy and average speed was $\rho = -0.17$ ($p = 0.67$), giving no evidence that a prompt's accuracy ranking predicts its speed ranking. In this data, prompt effectiveness did not transfer between the two task types, although with only nine prompt styles, only a strong relationship could have been detected.

A second pattern is that mean values for both task types are shaped heavily by a minority of outlier iterations. For the accuracy tasks, a majority of the median values tied while the means were more spread out. In the speed tasks, occasional scores of zero lowered the means of otherwise strong prompts. Before correction, prompt style had a significant effect on three of the five tasks (Prisoner Challenge, Message, and Number Partitioning); after Holm adjustment, on two (Table~\ref{tab:kw}). The size of prompt-style effects therefore appears to differ between tasks. On Migrations and Maximum Clique Size, where no significant effect was detected, the ranking of prompts by mean mostly reflects which prompts happened to have outlier runs, although with nine runs per prompt, smaller effects on these tasks cannot be ruled out.

\subsection{AI Performance Relative to Human Contestants}

There is ongoing debate over whether AI can now program as well as the best human programmers, or whether humans still hold an advantage. Competitive programming has become a common test of this question, and the evidence is mixed. OpenAI reported that its o3 model achieved gold-medal level performance at IOI 2024~\cite{openai2025competitive}. In contrast, a benchmark study done by Olympiad medalists found that frontier models solved none of the hard problems that human experts can solve, attributing high performance to implementation precision and tool augmentation rather than superior reasoning~\cite{zheng2025livecodebenchpro}.

\begin{figure}[t]
\centering
\includegraphics[width=\columnwidth]{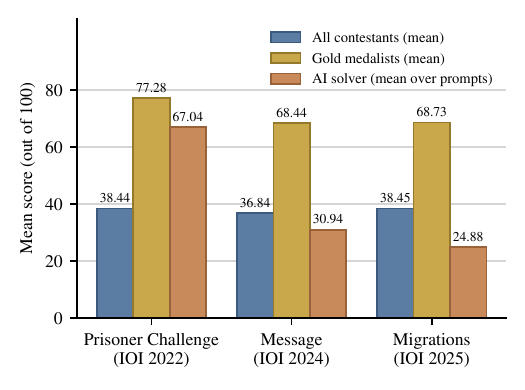}
\caption{AI performance relative to human contestants on the IOI accuracy tasks, scores are out of 100. Bars show the mean score for all contestants, the mean score of gold medalists, and the AI solver (averaged across all nine prompt styles). Contestant and gold-medalist scores were obtained from the official IOI statistics~\cite{ioi2022results,ioi2024results,ioi2025results}.}
\label{fig:humans}
\end{figure}

Read as descriptive context rather than a head-to-head test (Table~\ref{tab:humans}), the AI solver in our experiments scored below the contestant average on every task except Prisoner Challenge, where it outperformed the average contestant but still fell short of the gold medalists (Fig.~\ref{fig:humans}). While the AI solver was not strictly better than human contestants in this setting, it was not strictly worse either, though its advantage appeared on only one of the three tasks.

The AI's standing relative to humans changed sharply from task to task, while the average contestant's score barely changed. The average contestant scored between 36.84 and 38.45 on all three tasks, yet the AI's mean ranged from 24.88 on Migrations to 67.04 on Prisoner Challenge (Fig.~\ref{fig:humans}). On Prisoner Challenge, the AI scored 28.60 points above the average contestant and came within 10.24 points of the gold medalists; on Message and Migrations, it scored below the average contestant and less than half of the gold-medalist mean. The AI's scores also fell with the year of the task, from 2022 to 2024 to 2025. One possible explanation is exposure during training: all three tasks, and published solutions to them, predate GPT-5.5's documented knowledge cutoff of December 1, 2025~\cite{openai2026gpt55}, and older tasks have had longer for solutions and discussion to accumulate online. With only three tasks, however, this cannot be separated from other differences between the tasks. Either way, how the AI compares to humans depends heavily on which task is used, so a comparison based on one or two tasks could make it look much stronger or weaker than it is.

These comparisons are not head-to-head, as the AI solver and the human contestants worked under substantially different conditions (Table~\ref{tab:humans}). The AI solver was GPT-5.5 at medium reasoning effort with only single-shot generation, whereas the IOI contestants had five hours to solve three tasks and a fifty-submission limit per task. For each contestant, the task score is the sum of their best result on each subtask across all submissions, while the AI solver produced a single submission per iteration with no feedback. These scores were therefore obtained under different conditions and are best read as descriptive context rather than a head-to-head result; we return to this in the Future Work section.

\section{Limitations}

Each prompt style was represented by a single generated prompt per task, so the results compare these particular prompts rather than every possible prompt in each style; a different prompt in the same style could perform differently. The nine prompts for a task also differ in length and in some content: for example, the Maximum Clique Size prompts for Prompts 4 and 6 explicitly point the solver to the random structure of the input graph. Effects attributed to style may therefore partly reflect these differences. All three IOI tasks predate GPT-5.5's knowledge cutoff~\cite{openai2026gpt55}, so the accuracy scores may be affected by prior exposure. The speed score measures the largest input solved within the time limit, uses one test case per input size, and was measured on a single fanless laptop with a fixed grading order; the two speed tasks also used different compilers, although prompts were only ever compared within a task. Finally, with nine runs per prompt, the statistical tests can detect only fairly large effects, and the conclusions apply to one model and five tasks.

\section{Future Work}

As noted in the human comparison section, the evaluation was not head-to-head; evaluating AI performance with feedback, by allowing the solver to submit, observe its results, and resubmit, would give a more even comparison to human contestants. Using higher reasoning effort would likewise show how the solver performs at its best. Generating several independent prompts per style would show whether the differences belong to the styles or to the particular prompts, and grading the speed tasks in a randomized order with several test cases per input size would make the speed results more robust. Increasing the number of iterations per prompt could further clarify whether differences in the mean were caused by outlier iterations, give the statistical tests more power to detect smaller effects, and reduce the randomness that LLMs introduce into these comparisons. Similarly, increasing the number of tasks would help establish whether these results hold beyond the specific tasks studied here, and using tasks released after the model's training data was collected would test whether the AI's stronger result on the oldest task reflects prior exposure to it. Evaluating the same prompt styles across multiple models would test whether the observed effects are model-specific or general, and testing multiple programming languages would likewise show whether they are language-specific. Finally, combining prompt styles would test whether merging features from different styles outperforms any single style.

\section{Conclusion}

We investigated whether the way a prompt sent to an AI is structured affects the accuracy and speed of its output in competitive programming. Using one prompt per style on five tasks, we found that prompt style clearly affected performance on some tasks and showed no detectable effect on others, and that the most effective prompt depended on the task type (accuracy versus speed) and on the task itself. There is no single prompt style that consistently outperforms the others; the best choice is not stable. Many differences in mean scores were driven by a few outlier iterations, suggesting that the AI's output is less consistent than mean scores alone imply. When situated against human contestants on the IOI accuracy tasks, the AI was neither strictly better nor strictly worse in our setting, and its standing varied sharply from task to task. Taken together, and within the limits of this study, these results suggest that prompt style should be chosen to match the objective rather than treated as universal, and that evaluating prompting by a single metric can obscure how much its effectiveness varies.

% Generated by IEEEtran.bst, version: 1.14 (2015/08/26)

\appendices

\section{Prompt Style Descriptions}
\label{app:styles}

The nine prompt styles below were provided to the prompt-generation model, which produced one concrete prompt per style for each task. Appendix~\ref{app:prompts} lists the nine prompts generated for Number Partitioning; the full generated prompts for every task are available in the repository linked in Section~\ref{sec:data}.

\textbf{Prompt 1: Minimal lossless compression style.} Make the prompt as short as possible while keeping every solver-relevant detail.

\textbf{Prompt 2: Official statement style.} Rewrite in a formal, numbered, specification-like format with clear sections.

\textbf{Prompt 3: Symbol-definition style.} First define all symbols, variables, and entities clearly, then restate the task in a compact formal way.

\textbf{Prompt 4: Imperative directive style.} Phrase the prompt as direct commands to the solver, with strong instruction wording.

\textbf{Prompt 5: Structured checklist style.} Rewrite the task as a compact checklist of requirements and constraints.

\textbf{Prompt 6: Competitive benchmark style.} Present the task as a benchmark against other AI solvers. Use concise, competitive language that frames the objective as achieving the strongest possible solution under the task's requirements.

\textbf{Prompt 7: Educational teaching style.} Rewrite the task as if teaching the concept to a student encountering it for the first time. Introduce ideas gradually and prioritize intuition before formal details.

\textbf{Prompt 8: Plan-first algorithm design style.} Require the solver to first identify the core algorithm, data structures, invariants, and target complexity internally, then derive the implementation from that plan. Emphasize choosing the asymptotically strongest correct approach before coding.

\textbf{Prompt 9: Structured computational reasoning style.} Rewrite the task so that the solver is encouraged to internally represent the solution as a sequence of computational stages, decision rules, and iteration processes before implementation. Emphasize identifying the flow of computation and interactions between components prior to coding.

\section{Prompts Generated for Number Partitioning}
\label{app:prompts}

The nine prompts below were generated for Number Partitioning and sent to the solver as written, with the output instruction described in the Methodology appended to each.

\noindent\textbf{Prompt 1: Minimal lossless compression style}
\begin{quote}\small
Given N positive integers (line 1: N; next N lines: Ai), split all numbers into two groups minimizing |sum(group1)-sum(group2)|. Print that minimum difference. Example input: 4 1 5 7 9; output: 0. Tests run by increasing N and stop at first failure; TLE gives partial credit up to last solved N, WA/RTE gives zero. Time limit 1s. Use only the C++ standard library. Output only valid C++20 code, no explanation, no markdown.
\end{quote}

\noindent\textbf{Prompt 2: Official statement style}
\begin{quote}\small
Write a complete C++20 program using only the standard library.\\
1. Problem: You are given N positive integers. Assign each integer to exactly one of two groups. Minimize the absolute difference between the two group sums.\\
2. Input: The first line contains integer N. Each of the next N lines contains one positive integer Ai.\\
3. Output: Print one line containing the minimum possible absolute difference.\\
4. Example: input lines are 4, 1, 5, 7, 9; the correct output is 0.\\
5. Evaluation: Time limit is 1 second. Test cases are ordered by increasing N and run sequentially. If a Time Limit Exceeded occurs, evaluation stops and credit is based on the largest N solved before it. If a Wrong Answer or Run-Time Error occurs, evaluation stops and total credit is zero.\\
6. Requirements: the answer must be exact for every completed run, must compile under standard C++20, and must output only valid C++20 code with no explanation or code fences.
\end{quote}

\noindent\textbf{Prompt 3: Symbol-definition style}
\begin{quote}\small
Define N as the number of given values. Define A1, A2, ..., AN as positive integers read from input, one per line after N. Define a partition as a choice assigning every Ai to exactly one of two groups G0 and G1. Define S0 as the sum of values assigned to G0, S1 as the sum of values assigned to G1, and D = |S0 - S1|. Your program must compute min D over all possible partitions and print that value on one line. Input format: line 1 contains N; lines 2 through N+1 contain Ai. Example: for N=4 and values 1,5,7,9, the output is 0. Time limit: 1 second. Test cases are processed in increasing N; TLE stops scoring after the last successful N, while WA or RTE makes the total score zero. Implement an exact, robust solution in standard C++20 using only the C++ standard library. Output only valid C++20 code, with no prose and no markdown.
\end{quote}

\noindent\textbf{Prompt 4: Imperative directive style}
\begin{quote}\small
Read N. Read the next N positive integers. Place every integer into exactly one of two groups. Minimize the absolute difference between the two group sums. Print only that minimum difference followed by a newline. Preserve exact correctness; never use an approximation that can change the answer. Account for the evaluation rule: cases are ordered by increasing N, TLE stops with partial progress, but any WA or RTE gives zero. Optimize for the largest N possible within 1 second while remaining exact. Use only the C++ standard library and compile as standard C++20. For the sample values N=4 with 1, 5, 7, 9, the output is 0. Return only the C++20 source code; do not include explanations, comments outside code requirements, markdown, or code fences.
\end{quote}

\noindent\textbf{Prompt 5: Structured checklist style}
\begin{quote}\small
Checklist for the required C++20 program:\\
- Input: first line is N; each of the next N lines is a positive integer Ai.\\
- Operation: assign each Ai to one of two groups, using every value exactly once.\\
- Objective: minimize the absolute difference between the sums of the two groups.\\
- Output: one line containing the exact minimum possible difference.\\
- Example: input values 4, then 1, 5, 7, 9; output 0.\\
- Environment: standard C++20, standard library only, 1 second time limit.\\
- Evaluation behavior: tests are sorted by increasing N and stop at the first failure; TLE gives credit only through the previous successful N; WA or RTE gives total score zero.\\
- Solver output restriction: output only valid C++20 source code, no markdown, no explanation, no code fences.
\end{quote}

\noindent\textbf{Prompt 6: Competitive benchmark style}
\begin{quote}\small
You are competing against other solvers on an exact two-way partition benchmark. Given N positive integers from stdin, split them into two groups so the absolute difference of group sums is as small as possible, then print that minimum difference. Input is N on the first line followed by N lines containing Ai. Example: 4 / 1 / 5 / 7 / 9 must print 0. Time limit is 1 second, C++20 standard library only. Tests are ordered by increasing N; a TLE stops the run and earns credit up to the largest solved N, while any WA or RTE scores zero, so exactness and robustness are mandatory. Produce the strongest exact implementation you can for large N under these rules. Output only valid C++20 code, with no explanation or markdown.
\end{quote}

\noindent\textbf{Prompt 7: Educational teaching style}
\begin{quote}\small
Imagine you have a list of positive integers and two empty boxes. For each number, you must choose exactly one box. After all numbers are placed, each box has a sum. The goal is to make those two sums as close as possible, so you must output the smallest possible absolute difference between them. Formally, read N from the first line, then read N positive integers Ai, one per following line. Compute the exact minimum of |sum(box 1) - sum(box 2)| over all assignments. Print that value on one line. For example, with N=4 and numbers 1, 5, 7, 9, the numbers can be split into equal total sums, so the output is 0. Use standard C++20 and only the standard library. The time limit is 1 second. Test cases are tried in increasing N; if your program times out, scoring stops at the last N it solved, but if it gives a wrong answer or crashes, the total score is zero. Therefore, prefer a correct exact method and make it as efficient as possible. Output only the C++20 program, no explanatory text and no code fences.
\end{quote}

\noindent\textbf{Prompt 8: Plan-first algorithm design style}
\begin{quote}\small
Internally plan before coding. First identify the core exact algorithm for minimizing the difference between two subset sums, the data structures needed, the invariants that guarantee exactness, and the target time and memory complexity. Choose the asymptotically strongest correct approach you can implement in standard C++20 under a 1 second limit, considering that tests increase by N and that TLE only gives partial credit while WA or RTE gives zero. Then implement from that plan. Input: first line N, followed by N lines each containing a positive integer Ai. Output: the exact minimum possible value of |sum(group 1) - sum(group 2)| after assigning every Ai to exactly one of two groups. Example: N=4 with values 1, 5, 7, 9 outputs 0. Use only the C++ standard library. Your response must be only valid C++20 source code, with no markdown or explanation.
\end{quote}

\noindent\textbf{Prompt 9: Structured computational reasoning style}
\begin{quote}\small
Represent the computation internally as stages before implementing: parse N and the N positive integers; compute the total-sum information needed; explore or encode reachable choices of assigning numbers to two groups using exact decision rules; select the reachable split whose sums minimize the absolute difference; print that minimum. Maintain correctness invariants so every integer is assigned exactly once and every candidate difference considered by the algorithm corresponds to a real partition. Optimize the iteration flow and component interactions for the largest N possible in 1 second, because tests run in increasing N; a TLE stops with partial credit, but any wrong answer or run-time error yields zero. Input format is one integer N on the first line and one positive integer Ai on each of the next N lines. Output one line with the exact minimum absolute difference. Example: 4, then 1, 5, 7, 9 gives output 0. Compile as standard C++20 using only the standard library. Output only valid C++20 code, no prose, no markdown, no code fences.
\end{quote}


\begin{thebibliography}{10}
\providecommand{\url}[1]{#1}
\csname url@samestyle\endcsname
\providecommand{\newblock}{\relax}
\providecommand{\bibinfo}[2]{#2}
\providecommand{\BIBentrySTDinterwordspacing}{\spaceskip=0pt\relax}
\providecommand{\BIBentryALTinterwordstretchfactor}{4}
\providecommand{\BIBentryALTinterwordspacing}{\spaceskip=\fontdimen2\font plus
\BIBentryALTinterwordstretchfactor\fontdimen3\font minus
  \fontdimen4\font\relax}
\providecommand{\BIBforeignlanguage}[2]{{%
\expandafter\ifx\csname l@#1\endcsname\relax
\typeout{** WARNING: IEEEtran.bst: No hyphenation pattern has been}%
\typeout{** loaded for the language `#1'. Using the pattern for}%
\typeout{** the default language instead.}%
\else
\language=\csname l@#1\endcsname
\fi
#2}}
\providecommand{\BIBdecl}{\relax}
\BIBdecl

\bibitem{sclar2023quantifying}
\BIBentryALTinterwordspacing
M.~Sclar, Y.~Choi, Y.~Tsvetkov, and A.~Suhr, ``Quantifying language models'
  sensitivity to spurious features in prompt design or: How {I} learned to
  start worrying about prompt formatting,'' arXiv preprint arXiv:2310.11324,
  2023. [Online]. Available: \url{https://arxiv.org/abs/2310.11324}
\BIBentrySTDinterwordspacing

\bibitem{santana2025prompting}
\BIBentryALTinterwordspacing
E.~G. {Santana Jr} \emph{et~al.}, ``Which prompting technique should {I} use?
  {An} empirical investigation of prompting techniques for software engineering
  tasks,'' arXiv preprint arXiv:2506.05614, 2025. [Online]. Available:
  \url{https://arxiv.org/abs/2506.05614}
\BIBentrySTDinterwordspacing

\bibitem{openai2025competitive}
\BIBentryALTinterwordspacing
{OpenAI}, ``Competitive programming with large reasoning models,'' arXiv
  preprint arXiv:2502.06807, 2025. [Online]. Available:
  \url{https://arxiv.org/abs/2502.06807}
\BIBentrySTDinterwordspacing

\bibitem{zheng2025livecodebenchpro}
\BIBentryALTinterwordspacing
Z.~Zheng \emph{et~al.}, ``{LiveCodeBench Pro}: How do olympiad medalists judge
  {LLMs} in competitive programming?'' arXiv preprint arXiv:2506.11928, 2025.
  [Online]. Available: \url{https://arxiv.org/abs/2506.11928}
\BIBentrySTDinterwordspacing

\bibitem{padiyar2025analyzing}
\BIBentryALTinterwordspacing
S.~Padiyar, A.~Kurchania, D.~Singh, and M.~Boicu, ``Analyzing prompt
  engineering for code generation accuracy with o4-mini on {CodeForces}
  problems,'' \emph{Journal of Student-Scientists' Research}, vol.~7, 2025.
  [Online]. Available: \url{https://journals.gmu.edu/jssr/article/view/5156}
\BIBentrySTDinterwordspacing

\bibitem{ioi}
\BIBentryALTinterwordspacing
{International Olympiad in Informatics (IOI)}, ``Official website.'' [Online].
  Available: \url{https://ioinformatics.org}
\BIBentrySTDinterwordspacing

\bibitem{ioi2022prisoner}
\BIBentryALTinterwordspacing
{International Olympiad in Informatics}, ``{IOI} 2022 --- {Prisoner Challenge}
  problem statement,'' 2022. [Online]. Available:
  \url{https://ioinformatics.org/files/ioi2022problem2.pdf}
\BIBentrySTDinterwordspacing

\bibitem{ioi2024message}
\BIBentryALTinterwordspacing
------, ``{IOI} 2024 --- {Message} problem statement,'' 2024. [Online].
  Available: \url{https://ioinformatics.org/files/ioi2024problem2.pdf}
\BIBentrySTDinterwordspacing

\bibitem{ioi2025migrations}
\BIBentryALTinterwordspacing
------, ``{IOI} 2025 --- {Migrations} problem statement,'' 2025. [Online].
  Available: \url{https://ioinformatics.org/files/ioi2025problem5.pdf}
\BIBentrySTDinterwordspacing

\bibitem{garey1979}
M.~R. Garey and D.~S. Johnson, \emph{Computers and Intractability: A Guide to
  the Theory of {NP}-Completeness}.\hskip 1em plus 0.5em minus 0.4em\relax San
  Francisco, CA, USA: W. H. Freeman, 1979.

\bibitem{li2022alphacode}
\BIBentryALTinterwordspacing
Y.~Li \emph{et~al.}, ``Competition-level code generation with {AlphaCode},''
  arXiv preprint arXiv:2203.07814, 2022. [Online]. Available:
  \url{https://arxiv.org/abs/2203.07814}
\BIBentrySTDinterwordspacing

\bibitem{zou2025liveoibench}
\BIBentryALTinterwordspacing
K.~Zou, A.~Xiong, Y.~Zhang, X.~F. Zhang, Y.~Ren, J.~Yang, A.~Lee, S.~Bhushan,
  and L.~Wang, ``{LiveOIBench}: Can large language models outperform human
  contestants in informatics olympiads?'' arXiv preprint arXiv:2510.09595,
  2025. [Online]. Available: \url{https://arxiv.org/abs/2510.09595}
\BIBentrySTDinterwordspacing

\bibitem{jiang2023selfplanning}
\BIBentryALTinterwordspacing
X.~Jiang, Y.~Dong, L.~Wang, Z.~Fang, Q.~Shang, G.~Li, Z.~Jin, and W.~Jiao,
  ``Self-planning code generation with large language models,'' arXiv preprint
  arXiv:2303.06689, 2023. [Online]. Available:
  \url{https://arxiv.org/abs/2303.06689}
\BIBentrySTDinterwordspacing

\bibitem{li2023scot}
\BIBentryALTinterwordspacing
J.~Li, G.~Li, Y.~Li, and Z.~Jin, ``Structured chain-of-thought prompting for
  code generation,'' arXiv preprint arXiv:2305.06599, 2023. [Online].
  Available: \url{https://arxiv.org/abs/2305.06599}
\BIBentrySTDinterwordspacing

\bibitem{qiu2025efficient}
\BIBentryALTinterwordspacing
R.~Qiu, W.~W. Zeng, J.~Ezick, C.~Lott, and H.~Tong, ``How efficient is
  {LLM}-generated code? {A} rigorous \& high-standard benchmark,'' in
  \emph{Proc. Int. Conf. on Learning Representations (ICLR)}, 2025, arXiv
  preprint arXiv:2406.06647. [Online]. Available:
  \url{https://arxiv.org/abs/2406.06647}
\BIBentrySTDinterwordspacing

\bibitem{niu2024evaluating}
\BIBentryALTinterwordspacing
C.~Niu, T.~Zhang, C.~Li, B.~Luo, and V.~Ng, ``On evaluating the efficiency of
  source code generated by {LLMs},'' in \emph{Proc. 1st AI Foundation Models
  and Software Engineering (FORGE)}, 2024, arXiv preprint arXiv:2404.06041.
  [Online]. Available: \url{https://arxiv.org/abs/2404.06041}
\BIBentrySTDinterwordspacing

\bibitem{ojuz}
\BIBentryALTinterwordspacing
{oj.uz}, ``Online judge platform.'' [Online]. Available: \url{https://oj.uz}
\BIBentrySTDinterwordspacing

\bibitem{konc2007maxclique}
\BIBentryALTinterwordspacing
J.~Konc and D.~Jane{\v{z}}i{\v{c}}, ``An improved branch and bound algorithm
  for the maximum clique problem,'' \emph{MATCH Communications in Mathematical
  and in Computer Chemistry}, vol.~58, no.~3, pp. 569--590, 2007. [Online].
  Available: \url{http://insilab.org/articles/match2007.pdf}
\BIBentrySTDinterwordspacing

\bibitem{kruskal1952ranks}
\BIBentryALTinterwordspacing
W.~H. Kruskal and W.~A. Wallis, ``Use of ranks in one-criterion variance
  analysis,'' \emph{Journal of the American Statistical Association}, vol.~47,
  no. 260, pp. 583--621, 1952. [Online]. Available:
  \url{https://doi.org/10.1080/01621459.1952.10483441}
\BIBentrySTDinterwordspacing

\bibitem{spearman1904proof}
\BIBentryALTinterwordspacing
C.~Spearman, ``The proof and measurement of association between two things,''
  \emph{The American Journal of Psychology}, vol.~15, no.~1, pp. 72--101, 1904.
  [Online]. Available: \url{https://doi.org/10.2307/1412159}
\BIBentrySTDinterwordspacing

\bibitem{virtanen2020scipy}
\BIBentryALTinterwordspacing
P.~Virtanen \emph{et~al.}, ``{SciPy} 1.0: Fundamental algorithms for scientific
  computing in {Python},'' \emph{Nature Methods}, vol.~17, pp. 261--272, 2020.
  [Online]. Available: \url{https://doi.org/10.1038/s41592-019-0686-2}
\BIBentrySTDinterwordspacing

\bibitem{holm1979}
S.~Holm, ``A simple sequentially rejective multiple test procedure,''
  \emph{Scandinavian Journal of Statistics}, vol.~6, no.~2, pp. 65--70, 1979.

\bibitem{ioi2022results}
\BIBentryALTinterwordspacing
{International Olympiad in Informatics}, ``{IOI} 2022 results,'' IOI
  Statistics. [Online]. Available:
  \url{https://stats.ioinformatics.org/results/2022}
\BIBentrySTDinterwordspacing

\bibitem{ioi2024results}
\BIBentryALTinterwordspacing
------, ``{IOI} 2024 results,'' IOI Statistics. [Online]. Available:
  \url{https://stats.ioinformatics.org/results/2024}
\BIBentrySTDinterwordspacing

\bibitem{ioi2025results}
\BIBentryALTinterwordspacing
------, ``{IOI} 2025 results,'' IOI Statistics. [Online]. Available:
  \url{https://stats.ioinformatics.org/results/2025}
\BIBentrySTDinterwordspacing

\bibitem{openai2026gpt55}
\BIBentryALTinterwordspacing
{OpenAI}, ``{GPT-5.5} model,'' OpenAI API documentation, 2026. [Online].
  Available: \url{https://developers.openai.com/api/docs/models/gpt-5.5}
\BIBentrySTDinterwordspacing

\end{thebibliography}
\end{document}